# Arc-to-line frame registration method for ultrasound and photoacoustic image-guided intraoperative robot-assisted laparoscopic prostatectomy

Hyunwoo Song[1,3], Shuojue Yang[3], Zijian Wu[3], Hamid Moradi[4], Russell H. Taylor[1,3], Jin U. Kang[2,3], Septimiu E. Salcudean[4], Emad M. Boctor[1,3]

[1] Department of Computer Science, Whiting School of Engineering, the Johns Hopkins University, Baltimore, MD, USA, 21218
[2] Department of Electrical and Computer Engineering, Whiting School of Engineering, the Johns Hopkins University, Baltimore, MD, USA, 21218
[3] Laboratory for Computational Sensing and Robotics, the Johns Hopkins University, Baltimore, MD, USA, 21218
[4] Department of Electrical and Computer Engineering, the University of British Columbia, Vancouver, BC V6T 1Z4, Canada

**Abstract**
**Purpose:** To achieve effective robot-assisted laparoscopic prostatectomy, the integration of transrectal ultrasound (TRUS) imaging system which is the most widely used imaging modality in prostate imaging is essential. However, manual manipulation of the ultrasound transducer during the procedure will significantly interfere with the surgery. Therefore, we propose an image co-registration algorithm based on a photoacoustic marker (PM) method, where the ultrasound/photoacoustic (US/PA) images can be registered to the endoscopic camera images to ultimately enable the TRUS transducer to automatically track the surgical instrument.
**Methods**: An optimization-based algorithm is proposed to co-register the images from the two different imaging modalities. The principle of light propagation and an uncertainty in PM detection were assumed in this algorithm to improve the stability and accuracy of the algorithm. The algorithm is validated using the previously developed US/PA image-guided system with a da Vinci surgical robot.
**Results**: The target-registration-error (TRE) is measured to evaluate the proposed algorithm. In both simulation and experimental demonstration, the proposed algorithm achieved a sub-centimeter accuracy which is acceptable in practical clinics (i.e., $1.15 \pm 0.29$ mm from the experimental evaluation). The result is also comparable with our previous approach (i.e., $1.05 \pm 0.37$ mm), and the proposed method can be implemented with a normal white light stereo camera and doesn't require highly accurate localization of the PM.
**Conclusion**: The proposed arc-to-line based frame registration algorithm that transform between the camera and the TRUS imaging frames can allow simple yet efficient integration of commercial US/PA imaging system into the laparoscopic surgical setting, and contribute to the image-guided surgical intervention applications.

**Acknowledgments:**
We would like to acknowledge to our sponsors and funding agencies: Funding from the National Science Foundation Career Award 1653322, National Institute of Health R01-CA134675, the Johns Hopkins University internal funds, Canadian Institutes of Health Research, CA Laszlo Chair in Biomedical Engineering held by Professor Salcudean, and Intuitive Surgical for the equipment support.

**Funding:**
This work was supported in part by NSF Career Award 1653322, NIH R01-CA134675, the Johns Hopkins University internal funds, Canadian Institutes of Health Research (CIHR), CA Laszlo Chair in Biomedical Engineering held by Professor Salcudean, and an equipment loan from Intuitive Surgical, Inc.

**Declaration:**
The authors have no conflict of interest.

## 1. Introduction

Prostatectomy is one of the most common surgical procedures to treat prostate cancer, where the entire prostate gland is physically removed. Over the decades, several different surgical approaches such as radical prostatectomy and the procedure approached through perineal region have been accepted [1, 2]. In particular, robot-assisted laparoscopic prostatectomy (RALP), in which the surgeon conducts the surgery with a teleoperated surgical robot (e.g., da Vinci surgical robot). More than 80% of the entire radical prostatectomy in the United States has been performed using da Vinci robots to shortened recovery time and reduced post-operative complications [3–5]. Despite the fact that RALP procedure has improved the surgical efficacy and accuracy, the surgeon perceives the surgical scene by solely relying on the endoscopic camera view. Although the stereoscopic endoscopic imaging can enable the surgeon to visually interpret the surgical procedure some critical information such as underlying tissue structure or molecular characteristics (e.g., prostate cancer, peripheral nerve network) are not available. In other words, the endoscopic camera view cannot distinguish between healthy and cancerous tissue, nor can it localize the nerve structures surrounding the prostate gland. Therefore, there has been an emerging demand for additional imaging modalities that can provide advanced sensing capability [6–9].

Transrectal ultrasound (TRUS) imaging has been used for prostate imaging as it can provide non-invasive and real-time imaging of the prostate with high spatial resolution [10–12]. Moreover, photoacoustic (PA) imaging can be obtained using TRUS without alternating the imaging setup, which can deliver molecular contrast of the target tissue. Several studies have reported that the intra-operative TRUS+PA imaging can improve the surgical outcome by providing additional tissue information [13–16]. However, the effective usage of TRUS+PA imaging during the surgery is challenging mainly due to the fact that ultrasonic scanning is usually conducted by manual manipulation of the ultrasound transducer. Such scanning procedure will significantly interfere with the surgical workflow since the surgeon or sonographer needs to find the desired imaging plane.

Thus, it is crucial to effectively integrate the TRUS+PA imaging system into the current surgical environment. In this regard, there have been multiple endeavors to implement the TRUS imaging system with RALP procedure. Mohareri, et al. [10] developed a real-time US image guidance system during RALP procedure, where a robotized TRUS transducer automatically rotates to track the da Vinci surgical instrument. Based on this, real-time PA imaging was also developed by attaching the optical source to the surgical instrument [16]. However, these techniques have limited accuracy and applicability as the registration is based on da Vinci kinematics and acoustic detection of the surgical instrument by physical contact to the soft tissue.

To overcome this, a photoacoustic marker (PM) technique was recently introduced [17] as a novel tool for frame co-registration between the camera and the ultrasound imaging frames, where multiple laser excitations can be detected simultaneously by the camera and the ultrasound transducer in the form of laser spots and photoacoustic sources, respectively (Fig. 1). Hence, the acquired spots in both frames are used to calculate the coordinate transformation between the frames. Based on the proof-of-concept validation, this technique was recently implemented with a TRUS imaging system equipped with a compact near-infrared (NIR) pulsed laser diode (PLD), and demonstrated *ex vivo* in a da Vinci SI surgical robot environment [18]. The study showed that the system can be easily implemented to the surgery by minimizing the interference during the procedure, where only a single optical fiber is inserted into the surgical area (i.e., abdomen). Furthermore, the PM based frame registration has achieved sub-centimeter accuracy (e.g., 1.05±0.37mm in target registration error (TRE)) that can be practically accepted for the imaging [19, 20]. However, the current technique still suffers from several limitations: (1) The low-average power laser source can be a hurdle for endoscopic camera to detect the laser spot during the in vivo environment with the strong tissue and blood absorption. In addition, the conventional approach requires the fluorescence (FL) camera to detect the FL spot as the PM. Hence, the user needs to switch the laser on and off during the surgery. One might suggest that the laser spot can be segmented by adding a visible laser source to the existing laser path, but visible light spot can distract the surgeon's sight which disturbs the surgery. (2) The accuracy of the frame transformation dependent on the detection accuracies of laser spot and photoacoustic source. In particular, the TRUS transducer rotates to find the acoustic signal until it obtains the optimal position of the photoacoustic source. Note that such a searching process via manual manipulation of the TRUS transducer or algorithmic approach can be less accurate and time-consuming, respectively.

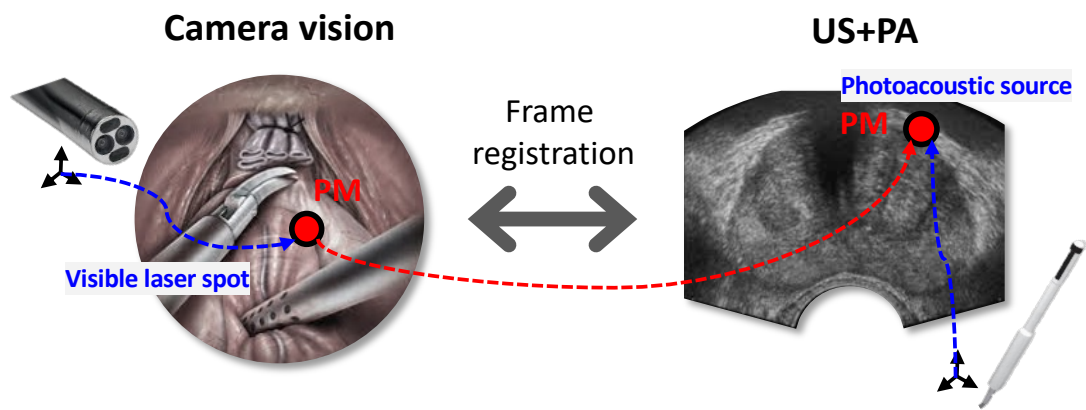

**Fig. 1. Conceptual illustration of the photoacoustic marker (PM) based frame registration method.** The camera image and the ultrasound + photoacoustic image are co-registered with the PM which can detected by both camera and the ultrasound sensor as forms of the visible laser spot and photoacoustic source, respectively.

To address these limitations, we propose a novel algorithm based on arc-to-line frame registration between the camera and the ultrasound imaging frames. The proposed algorithm, rather than directly localizing the laser spot, leverages the principle of the laser beam, to parameterize the beam into a line vector. In addition, we have imposed a rotation uncertainty to the photoacoustic source detection by the TRUS transducer, under the assumption that the detected source position is not accurate but located within the arc having a radius equivalent to the distance between the sensor and the source. Based on these parameterizations, an iterative nonlinear optimization algorithm solves the frame registration by estimating the laser spot and corresponding acoustic source in each frame.

In the following sections, the line-to-arc based frame registration algorithm will be described in detail with its implementation into the current system that we developed. Also, quantitative evaluation and analysis will be discussed based on our computer simulation and experimental validation. In this work, we have focused on developing and validating the proposed algorithm, and its real-time demonstration will be further introduced and discussed in future work.

## 2. Methods
### 2.1. Line-to-arc based frame registration

The ultimate objective of the proposed algorithm is to co-register the two image frames, from the endoscopic camera and the TRUS transducer (Fig. 2). Note that the algorithm calculates the frame transformation by using the PMs estimated from each frame without direct localization from the camera view, or precise searching by the TRUS transducer. Here, the algorithm is valid when the following conditions are satisfied: (1) the laser beam is straight; (2) The acoustic wave from the out-of-plane source can still be detected when the source is within the transducer's sensitivity range (i.e., slice thickness, Fig. 3). Assuming

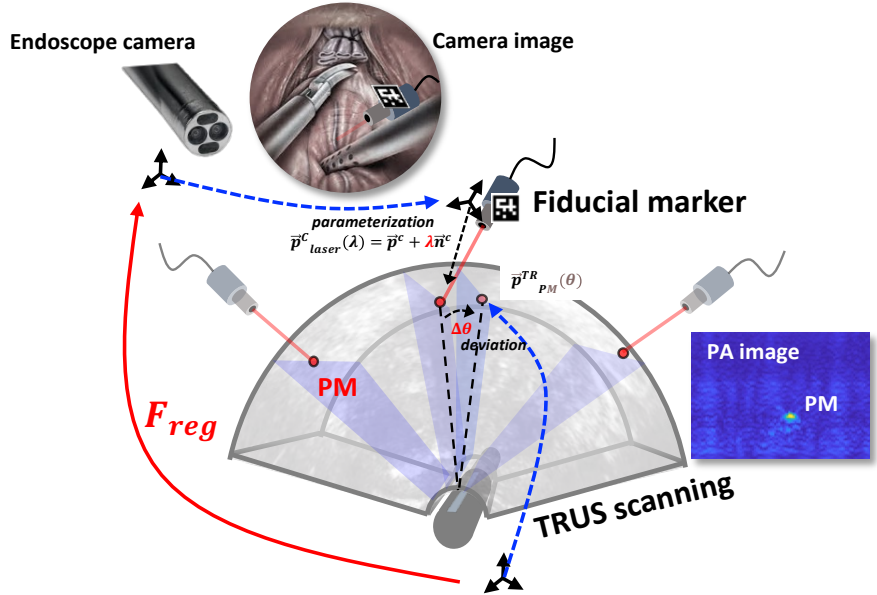

**Fig. 2. Conceptual illustration of the proposed arc-to-line registration method.** The endoscopic camera detects the fiducial marker attached to the optical fiber to estimate the excited laser beam. The TRUS transducer collects the corresponding PM assuming that the detected position is not perfectly accurate. $F_{reg}$: the frame transformation between the camera and the TRUS transducer; $\vec{p}^C_{laser}$: the parameterized laser beam; $\vec{p}^{TR}_{PM}$: the position of the detected PM in the TRUS transducer frame

in an isotropic medium, the laser beam is shown to propagate in a straight line. Thus, one can parameterize the excited laser beam as a 3-dimensional (3-D) line vector,

$$\vec{p}^C_{laser_i}(\lambda_i) = \vec{p}^C_i + \lambda_i \vec{n}^C_i \tag{1}$$

where $\vec{p}^C_i$, and $\vec{n}^C_i$ indicate any point lying on the laser line (e.g., excitation point), and the orientation of the laser (i.e., unit vector) at *i*-th position, respectively.

Here, the position of the PM on the tissue surface can be estimated by calculating the appropriate $\lambda$. Note that the parameterized line vector is based on the endoscopic camera frame. Here, a fiducial marker is attached to the optical fiber head where its pose can be estimated by the endoscopic camera. Based on the estimated pose of the optical fiber, the system calculates the orientation of the excited laser (i.e., $\vec{n}^C_i$), and hence parameterizes the laser pathway. The detailed process for the calibration of the laser parameterization will be discussed in the following section.

As introduced earlier, the PM can be also detected by the ultrasound transducer as form of a photoacoustic source. Here, the acoustic wave is generated at the PM position by the photoacoustic effect, propagates through the tissue, and can be received by the TRUS transducer. The position of the corresponding *i*-th PM with respect to the TRUS transducer can be expressed as

$$\vec{p}^{TR}_{PM_i} = [x_{PM_i}, y_{PM_i}, z_{PM_i}]^T \tag{2}$$

To acquire the PM position in TRUS transducer frame, the transducer rotates along the longitudinal axis until it detects an acoustic signal from the PM in the reconstructed 2D PA image. Note that the TRUS transducer frame is stationary which does not move with the rotation (Fig 3). In particular, if the transducer detects the *i*-th PM in the PA image at scanning angle $\theta_i$ (Fig. 3), the position of PM (i.e., $\vec{p}^{TR}_{PM_i}$) can be converted as a function of $\theta$ such that,

$$\vec{p}^{TR}_{PM_i}(\theta_i) = [x_{e_i}, y_{e_i}, z_{e_i}]^T + [0, r_{PM_i} \sin\theta_i, r_{PM_i} \cos\theta_i]^T \tag{3}$$

where $[x_{e_i}, y_{e_i}, z_{e_i}]^T$ is the 3-D position of the transducer element at scanning angle $\theta_i$, $r_{PM_i}$ indicates the distance between the transducer element and the photoacoustic source (i.e., PM). Here, $[x_{e_i}, y_{e_i}, z_{e_i}]^T$ is defined with the transducer specification and current orientation angle $\theta_i$. Given the number of transducer element ($N_{ele}$), interval between the elements ($d_{pitch}$), and the radius of the transducer ($r_{TRUS}$), the position of *i*-th transducer element $[x_{e_i}, y_{e_i}, z_{e_i}]^T$ can be defined as

$$x_{e_i} = [-0.5 * (N_{ele} - 1) + i] * d_{pitch} \tag{4-1}$$

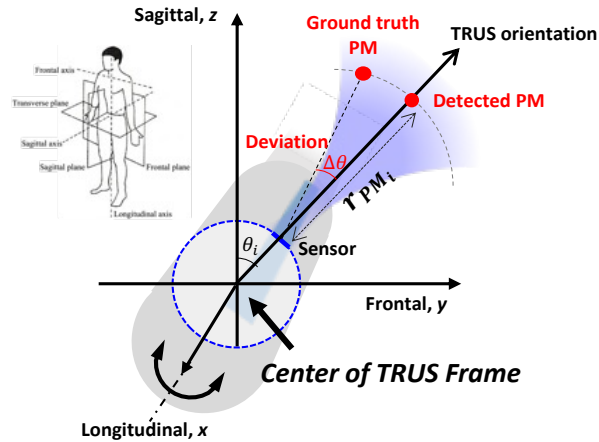

**Fig. 3. Geometrical description of the TRUS transducer frame, and the slice thickness of the ultrasound transducer.** The photoacoustic signal from the PM can be received by the TRUS transducer when the PM is located within the slice thickness of the transducer. The blue shaded area indicates the acoustic sensitivity of the sensor in the transverse plane.

$$y_{e_i} = \sin\theta_i * r_{\text{TRUS}} \qquad (4\text{-}2)$$
$$z_{e_i} = \cos\theta_i * r_{\text{TRUS}} \qquad (4\text{-}3)$$

where $i = 0, 1, 2, \cdots, N_{ele}\text{-}1$

However, it is important to remember that the acoustic sensor has a certain sensitivity range, the acoustic signal can still be received when it is within the slice thickness of the transducer, even if the PM is not aligned with the orientation of the TRUS transducer. In other words, the position of the actual photoacoustic source can be deviated by an unknown angle $\Delta\theta$ from the current scanning orientation (Fig. 3). Note that the deviated source is located within the arc with a radius of $r_{PM}$, since the signal is detected at equivalent time where the acoustic wave propagates from the source to the transducer element. As a result, the revised expression for the position of photoacoustic source is

$$\vec{p}^{*TR}_{PM_i}(\theta'_i) = [x_{e_i}, y_{e_i}, z_{e_i}]^T + [0, r_{PM_i}\sin(\theta'_i), r_{PM_i}\cos(\theta'_i)]^T \qquad (5)$$

where $\theta'_i = \theta_i + \Delta\theta_i$.

Thus, the frame transformation between the endoscopic camera and the TRUS transducer ($F_{reg}$) can be derived by solving a nonlinear cost function consists of Eq. 1 and Eq. 3,

$$J(F_{reg}, \vec{\lambda}, \vec{\theta'}) = \sum_i \left\| F_{reg} \cdot \vec{p}^{C}_{laser_i}(\lambda_i) - \vec{p}^{*TR}_{PM_i}(\theta'_i) \right\|_2 \qquad (6)$$

where $\vec{\lambda} = [\lambda_0, \lambda_1, \cdots, \lambda_i]^T$, and $\vec{\theta'} = [\theta'_0, \theta'_1, \cdots, \theta'_i]^T$.

Here, we propose to solve the cost function in an iterative manner:

(1) First of all, use gradient descent method to solve $F_{reg}$ and $\vec{\lambda}$ that minimizes the cost function (i.e., Eq. 6) such that

$$F^*_{reg}, \vec{\lambda}^* = \operatorname{argmin}_{F_{reg}, \vec{\lambda}} \left\{ \sum_i \left\| F_{reg} \cdot \vec{p}^{C}_{laser_i}(\lambda_i) - \vec{p}^{*TR}_{PM_i}(\theta'_i) \right\|_2 \right\} \qquad (7)$$

Note that we leave the PMs in the TRUS frame as deviated ones.

(2) Next, we compensate the deviated PM in the TRUS frame to estimate the actual scanning angle ($\vec{\theta^*}$) using gradient descent method, given the calculated $F^*_{reg}$ and $\vec{\lambda}^*$

$$\vec{\theta^*} = \operatorname{argmin}_{\vec{\theta'}} \left\{ \sum_i \left\| F^*_{reg} \cdot \vec{p}^{C}_{laser_i}(\lambda^*_i) - \vec{p}^{*TR}_{PM_i}(\theta'_i) \right\|_2 \right\} \qquad (8)$$

(3) From step (1) and (2), update the 3-D positions of PMs in both coordinates by defining the PMs from the parameterized lines and arcs, respectively. For example, The updated position of $i$-th PM in both frames are $\vec{p}^{C}_{laser_i}(\lambda^*_i) = \vec{p}_i + \lambda^*_i \vec{n}_i$ and $\vec{p}^{*TR}_{PM_i}(\theta^*_i) = [x_{e_i}, y_{e_i}, z_{e_i}]^T + [0, r_{PM_i}\sin(\theta^*_i), r_{PM_i}\cos(\theta^*_i)]^T$ relative to endoscopic camera frame and TRUS transducer frame, respectively.

(4) Iterate the steps (1) – (3) until the cost function converges.

Fig. 4 summarizes the entire workflow of calculating the frame registration between the endoscopic camera and the TRUS transducer. At first, the user acquires multiple datasets ($N_s$) composed of: (1) the pose of the attached fiducial marker in the camera frame (i.e., $F_{MC}^i$); and (2) the position of the corresponding PM in TRUS transducer frame (i.e., $\vec{p}_{PM_i}^{*TR}$). Note that a single laser source is used for generating the PM at different positions, and the TRUS transducer rotates to search for the PM. Here, we assume that the calibration between the fiducial marker and the optical fiber is already conducted. Consequently, the laser beam is parameterized with respect to the camera coordinate, and $F_{reg}$ is initialized. Afterwards, the proposed algorithm described above solves the cost function to calculate the final $F_{reg}$.

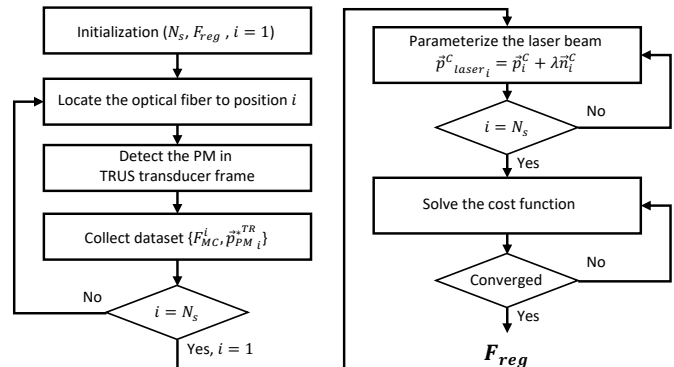

**Fig. 4. Workflow chart of the registration process.**

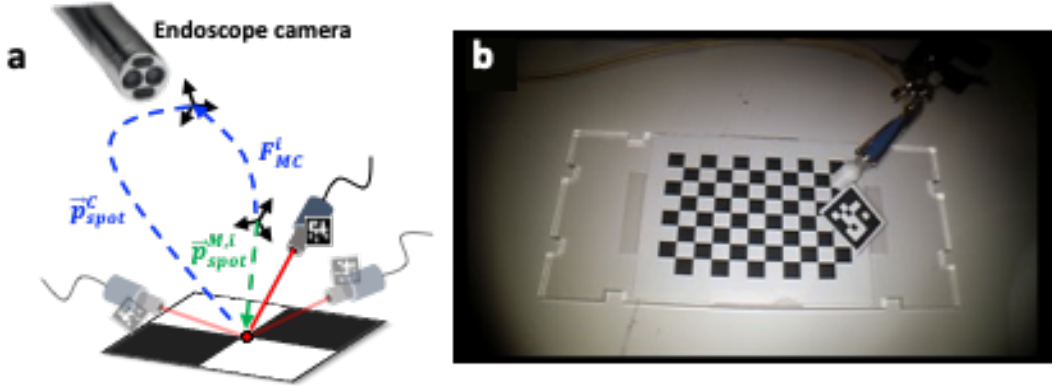

**Fig. 5. (a) The illustration of calibration process, (b) Calibration setup.**

### 2.2. Calibration

As mentioned in the section 2.1, the camera captures the pose of the fiducial marker attached to the optical fiber head. Consequently, the orientation of the excited laser is estimated by parameterizing the beam (i.e., Eq. 1). Here, a calibration process is conducted in a similar manner to pivot calibration, to associate the laser beam relative to the fiducial marker frame. Fig. 5 shows an illustration of the calibration process (Fig 5(a)), and experimental setup with da Vinci endoscope camera (Fig 5(b)). A simple checkerboard is located at the field-of-view (FOV) of the endoscopic camera which captures the corners of the checkerboard. One of the corners are selected and we denote the 3-D position of this point as $\vec{p}_{spot}^C$. Consequently, the optical fiber is rotated multiple times while aiming at the same corner (i.e., $\vec{p}_{spot}^C$). Here, the camera captures each pose of the fiducial marker ($F_{MC}^i$), and we can derive the following linear relationship.

$$\begin{bmatrix} \vdots \\ \vec{p}_{spot}^{M,i} \\ \vdots \end{bmatrix} = \begin{bmatrix} F_{MC}^0 & \vdots & \mathbf{0} \\ \mathbf{0} & F_{MC}^i & \mathbf{0} \\ \mathbf{0} & \vdots & F_{MC}^{N_s} \end{bmatrix} \begin{bmatrix} \vdots \\ \vec{p}_{spot}^C \\ \vdots \end{bmatrix} \tag{9}$$

where $\vec{p}_{spot}^{M,i}$ is the $i$-th position of the laser spot in local fiducial marker frame, $N_s$ is the number of acquisitions.

An important aspect is that the sampled points in the local fiducial marker frame are co-linear to each other. In other words, as the laser is excited from the optical fiber head which is rigidly fixed to the fiducial marker, every point (i.e., $\vec{p}_{spot}^{M,i}$, $i = 0, 1, 2, \ldots, N_s$) lies on the excited laser line in the fiducial marker frame. Therefore, a straight line path can be estimated with the points by using a line fitting algorithm such as singular value decomposition (SVD). By denoting the orientation of the fitted line as $\vec{n}_l^M$, and choosing one of the sampled points in the fiducial marker frame (e.g., $\vec{p}_{spot}^{M,0}$), the parameterized line vector with respect to the camera frame (i.e., Eq. 1) can be re-expressed as

$$\vec{p}_{laser_i}^C(\lambda_i) = F_{MC}^i \cdot \vec{p}_{spot}^{M,0} + \lambda_i F_{MC}^i \cdot \vec{n}_l^M \tag{10}$$

### 2.3. Tracking

Based on the calculated frame transformation $F_{reg}$, the TRUS transducer can track the PM when it is at the out-of-plane, by rotating to the right orientation. Here, remember that we require two constraints: (1) the fiducial marker can be localized by the camera; (2) The PM is located within the transducer's sensitivity range. When the two constraints are satisfied, the tracking angle can be calculated by first solving the cost function given the calculated $F_{reg}$:

$$\lambda^*, \theta^* = \mathrm{argmin}_{\lambda, \theta'} \left\{ \left\| F_{reg} \cdot \vec{p}^C_{laser}(\lambda) - \vec{p}^{*TR}_{PM}(\theta') \right\|_2 \right\} \tag{11}$$

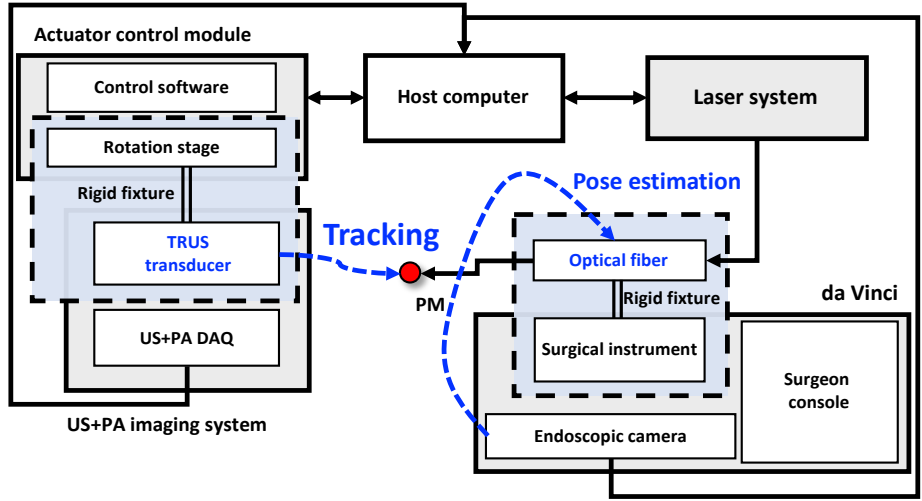

**Fig. 6. Overall system configuration.** The system is composed of the host computer, laser system, actuator control module, US+PA imaging system, and da Vinci SI surgical robot. The TRUS image and the endoscopic camera image are co-registered via the PM. The TRUS transducer tracks the PM generated by the laser excitation. The endoscopic camera detects the pose of the optical fiber.

Hence, the desired rotation angle to make the TRUS transducer to be in-plane to the PM can be calculated as $\Delta\theta = \theta^* - \theta$.

### 2.4. Performance evaluation via computer simulation

Computer simulation was conducted to evaluate the calibration and registration algorithm. First of all, calibration was evaluated by adding non-zero mean anisotropic gaussian noise (i.e., $\mu_{MC}$ = {0.1, 0.1, 0.1} mm, in x-, y-, z-direction, mean; $\sigma^{trans}_{MC}$ = {0.1, 0.1, 0.8}, $\sigma^{rot}_{MC}$ = {0.01, 0.01, 0.01} in x-, y-, z-direction, standard deviation of translation (mm) and rotation (radian), respectively) to the fiducial marker detection to reflect the systematic bias and detection noise in practical circumstances [21]. Additional noise was considered to represent the error when the laser is pointing at the same spot: a zero-mean gaussian noise where $\sigma_{aim}$ = {1, 1} mm in horizontal and vertical direction, respectively. Note that the gaussian kernel is in 2-dimension as the aiming object (i.e., checkerboard) is flat. By having the two noise components (i.e., fiducial marker localization and pointing the laser), the calibration performance was evaluated with increasing number of tracked poses: {5, 10, 20, 30, 40}.

In addition, the feasibility of the proposed registration algorithm was evaluated. Here, we assumed that the photoacoustic source is always generated, and the corresponding acoustic wave is received by the TRUS transducer without any external factors such as limited light absorbance, tissue inhomogeneity, or the sensitivity of the transducer (e.g., slice thickness). Here, same noise model was applied to the fiducial marker detection (i.e., $\mu_{MC}$ = {0.1, 0.1, 0.1}mm, in x-, y-, z-direction, mean; $\sigma^{trans}_{MC}$ = {0.1, 0.1, 0.8}, $\sigma^{rot}_{MC}$ = {0.01, 0.01, 0.01} in x-, y-, z-direction, standard deviation of translation (mm) and rotation (radian), respectively), and different degrees of deviation of detected PMs in the TRUS transducer frame are defined in order to investigate the tolerance of the proposed algorithm: {±5°, ±10°, ±15°, ±20°}.

### 2.5. Experimental demonstration in the da Vinci surgical robot environment

The proposed registration algorithm is evaluated not only in the computer simulation, but also demonstrated in our recently developed TRUS+PA image-guided system with a da Vinci SI surgical robot environment [18]. Note that we will skip the detailed description of the developed system, but focus on the revised or added components for this research. Fig. 6 shows the overall system configuration composed of the main host computer, da Vinci SI surgical robot system, an actuator control module, and a TRUS+PA imaging system. The system is developed on the Robot Operating System (ROS) platform, and each component communicate to each other in real-time.

In addition, the endoscopic stereo camera of the da Vinci system estimates the pose of the fiducial marker attached (ArUco marker with a size of 30×30 mm$^2$) to a custom-made mount that can be installed on the optical fiber body (QSP-785-4, QPhotonics LLC, Ann Arbor, MI, USA). Here, an open source library (i.e., OpenCV) based software is used for the pose estimation of the fiducial marker. The excited laser beam is propagated through a focusing lens (LB1406-B, Thorlabs, Newton, NJ, USA) to generate the PM on the target surface. A homogenous phantom made of plastisol material is used to maximize the light absorption to generate photoacoustic effect at the excited area. Consequently, the generated acoustic wave from the PM is searched and detected by a TRUS transducer (BPL 9-5/55, BK Medical, Peabody, MA, USA) actuated by a rotation stage (PRM1Z8, Thorlabs, Newton, NJ, USA). Here, the detection of the PM is conducted by manually rotating the TRUS transducer with 2° increment, and the searching of the PM stops once a signal is detected by the developed PM localization program (i.e., no further searching is conducted for accurate PM detection).

### 2.6. Quantitative evaluation

The performance of the proposed algorithm was quantitatively evaluated. In particular, we obtained 15 datasets (i.e., the pose of the fiducial marker and corresponding PM position in the camera and TRUS transducer frames, respectively) during the registration process, and arbitrarily chose 10 datasets from the entire dataset to calculate the frame transformation between the two frames. The accuracy of the registration result was evaluated by the TRE measurement using the remaining five datasets. The TRE is measured by the following equation:

$$\text{TRE} = \frac{1}{5}\sum_{i=1}^{5} \left\| F_{reg} \cdot \vec{p}^{\,C}_{laser_i}(\lambda_i) - \vec{p}^{\,*TR}_{PM_i}(\theta'_i) \right\|_2 \qquad (12)$$

The tracking error was also evaluated to quantify how accurate the TRUS transducer can rotate to localize the PM to provide the in-plane image of corresponding cross-section imaging plane. The tracking error is measured by taking the difference between the calculated angle and the desired angle rotation angle according to the true PM position.

## 3. Results

The proposed algorithm was demonstrated and validated in both computer simulation and experimental setup, and will be described in the following sections.

### 3.1. Evaluation in computer simulation

Based on the additive noise model described in Section 2.4, the calibration result showed that the residual errors were decreased as the number of tracked poses increases (Fig. 7a): {2.12 ± 0.81, 1.71 ± 0.42, 1.39 ± 0.41, 1.30 ± 0.38, 1.18 ± 0.27} mm when the number of datasets is {5, 10, 20, 30, 40}, respectively.

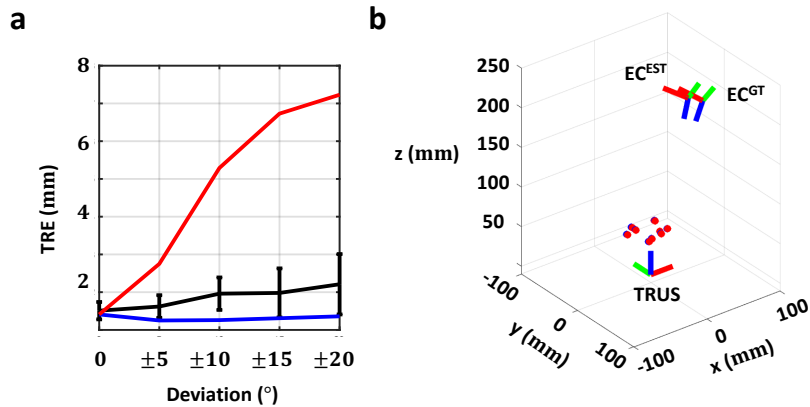

**Fig. 7. Performance evaluation of the proposed algorithm in computer simulation.** (a) TRE measurement. Black, blue, and red line indicates the TRE, upper, and lower bound, respectively. (b) The relative pose of the EC$^{EST}$ (Estimated endoscopic camera frame) and EC$^{GT}$ (Ground truth endoscopic camera frame).

Moreover, the proposed algorithm was validated whose TRE at each case was {2.21±1.08, 1.98±0.95, 1.96±0.98, 1.62±0.61, 1.51±0.61}-mm for {±20°, ±15°, ±10°, ±5°, 0°} of the deviation relative to the ground truth PM positions, respectively. Note that the zero deviation indicates that the detected PM is identical to the ground truth position. The blue and red line indicate the upper bound and the lower bound of the proposed algorithm, respectively. Fig. 7(b) shows the relative positions of the camera frame with respect to the TRUS transducer based on the recovered $F_{reg}$. Note that the frame of the camera becomes closer to the ground truth pose, and the measured TRE proportionally decreases as the deviation becomes smaller.

### 3.2. Experimental evaluation

The proposed algorithm is further validated experimentally in the previously described TRUS+PA image-guided system in a da Vinci environment [18]. In this section, the experimental evaluation will be described in each step (i.e., calibration, registration, and tracking).

#### 3.2.1. Calibration

The calibration was conducted by obtaining 12 different poses of the optical fiber (i.e., $F_{MC}^i$). As previously described, the centers of the 12 different laser beams were overlapped with the same corner on the checkerboard to ensure that the laser beam is aimed at the same point (i.e., $\vec{p}_{spot}^{C}$). The points in the local fiducial marker frame were derived according to Eq. (9), and line fitting was performed to estimate the orientation of the excited laser beam. Fig. 8 shows the converted positions of the laser spot with corresponding fitted line (i.e., Fig. 8). The average Euclidean distance between the point and the line was measured as the residual error (i.e., 1.55 ± 0.9 mm) which is within an acceptable range, and it is subject to change depending on the number of obtained tracked poses.

#### 3.2.2. Registration

Following the fiducial marker-to-laser calibration, the frame registration was performed to obtain $F_{reg}$ between the two frames (Fig. 9(a)). 15 datasets composed of the pose of the fiducial marker in the camera frame and corresponding PM positions in the TRUS transducer frame (i.e., $\{F_{MC}^i, \vec{p}_{*PM_i}^{TR}\}$) are collected for the calculation. Here, an important thing to remember is that the clinical TRUS transducer has certain slice thickness along the scanning direction, we have experimentally measured the sensitivity range by scanning from -35° to 35° to unveil how much deviated angle the TRUS transducer can receive the out-of-plane PM signal. Fig. 9(b) shows the sensitivity curve of the TRUS transducer at the scanned range (i.e., [-35°, 35°]), and the maximum angle that the transducer can receive the out-of-plane PM signal was ±6° at 30mm imaging depth. Note that the sensitivity range can be changed at deeper imaging depth. Next, a subset of the entire datasets was chosen to calculate the frame transformation $F_{reg}$. Fig. 9(c) shows the relative pose of the endoscopic camera with respect to the TRUS transducer based on the optimized $F_{reg}$. The TRE measurement revealed that the larger amount of collected pairs will contribute to more accurate calculation, as more noise components (e.g., fiducial marker detection error, and PM localization in the PA image, etc.) are suppressed during the optimization (Fig. 9(d)). In particular, four to ten datasets were arbitrarily chosen among the entire dataset for the calculation, and the averaged TRE values are: { 2.95 ± 0.87 mm, 2.23 ± 1.21 mm, 1.86 ± 0.52

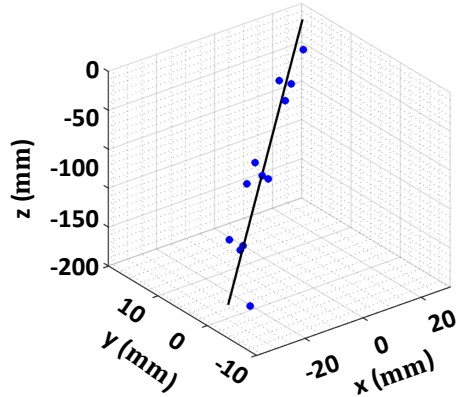

**Fig. 8. Visualization of the calibration result.** The blue dots and the black line indicate the converted positions of the laser spot and corresponding fitted line, respectively. The average residual error was 1.46mm

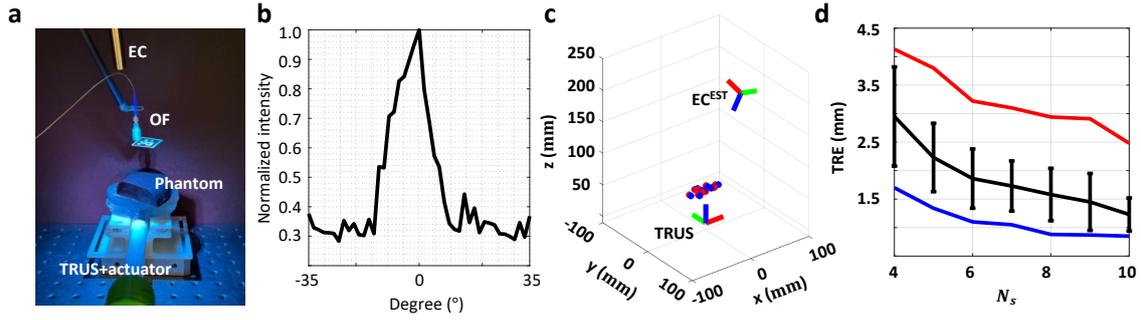

**Fig. 9. Experimental evaluation of the proposed algorithm.** (a) Experimental setup. EC: Endoscopic camera; OF: Optical fiber. (b) Acoustic intensity curve to represent the directional sensitivity of the transducer. (c) The relative pose of the estimated EC ($EC^{EST}$) with respect to the TRUS transducer. The blue and red dots indicate the estimated PMs in the TRUS and the camera frame, respectively. (d) The TRE measurement. Black, blue, and red line indicates the TRE, upper, and lower bound, respectively. $N_s$: number of pairs used in the registration

mm, 1.73 ± 0.44 mm, 1.58 ± 0.46 mm, 1.45 ± 0.5 mm, 1.15± 0.29 mm} for $N_s$ = 4, 5, 6, ⋯ , 10, respectively. Here, $N_s$ refers to the number of acquisitions. In addition, the proposed algorithm is quantitatively compared to the conventional PM approach, the method which directly detects the position of the PM both in the camera image and PA image, to investigate its clinical efficacy: 1.15 ± 0.29 mm vs. 1.05 ± 0.37 mm for the proposed method and the conventional approach, respectively. The two methods showed a comparable performance in its registration accuracy. Moreover, we have conducted a leave-one-out cross validation approach to statistically analyze the performance of the proposed algorithm given limited dataset ($N_s$ = 15). The cross validation shows that the average error is 2.06 ± 1.97 mm, implying the fact where the proposed algorithm is sensitive to the deviated PM with respect to the TRUS. Nonetheless, the proposed method has several advantages where it is not dependent on the FL detection of the PMs from the camera (i.e., more robust in detecting the fiducial marker than the FL spot), and the deviated PM can still be used in frame calculation. That being said, the proposed algorithm can significantly reduce the time for dataset acquisition during the registration, compared to that of previous method in [18]. Based on the sensitivity range of ±6° at 30mm imaging depth, and assuming that the PM can be resolved with an actuator that has 0.1° accuracy, the worst case of searching a field-of-view of 70° can be roughly estimated by 70°/12° + 12°/0.1° ≅ 126 times. On the other hand, our method can achieve the detection within 70°/12° ≅ 6 times. We have also measured the average convergence time of the proposed algorithm at each deviation: {50.6±49.38, 79.6±37.9, 81.84±19.61, 101.28±39.56, 125.84±35.4}-sec for {0°, 5°, 10°, 15°, 20°}. It implies that convergence time increases when the deviated PM is farther from the transducer. As a result, the proposed arc-to-line based registration algorithm provides the fact that it can play a better role for the US/PA image-guided surgical intervention during the RALP.

### 3.2.3. Tracking

Based on the calibrated laser detection and the co-registered frames between the camera and the TRUS transducer, the tracking of the PM by the TRUS transducer was tested. As the TRUS transducer is only rotating along the longitudinal axis (i.e., one degree-of-freedom), the tracking error was measured by calculating the difference between the angle derived from the optimization (i.e., Eq (10)) and the angle needs to be rotated in order to be aligned with the PM at the surface (i.e., in-plane). Based on the registration result, the evaluation was performed on the $F_{reg}$ which was calculated with 10 pairs: 1.06 ± 0.56°. Here, securing accurate tracking of the PM is important for precise imaging, and it becomes more significant when the imaging depth becomes deeper [22]. For example, tracking error of 1° at 30mm imaging depth will produce 0.5mm of imaging plane deviation in radial direction (i.e., 30mm * 1° * π / 180° ≅ 0.5mm), while it becomes larger when it becomes deeper (e.g., 60mm * 1° * π / 180° ≅ 1mm).

## 4. Discussion and conclusion

In this paper, we presented arc-to-line registration based frame transformation algorithm for registering the TRUS+PA image to the endoscopic camera during robot-assisted laparoscopic prostatectomy. In particular, the algorithm was experimentally evaluated in a widely used surgical robot environment, the da Vinci SI system, and provided stable registration performance with sub-centimeter accuracy. It was also comparatively evaluated with the previously developed PM technique, and further provided better applicability for the clinics by providing not only comparable accuracy, but also secures more robust fiducial marker detection from the camera and improved accuracy in detecting the PM in the US+PA image. Besides, the study does have some limitations that can be addressed by future work.

First of all, we will miniaturize the fiducial marker attached to the optical fiber head. Currently, the size of the marker is 30×30mm$^2$, which is relatively large in laparoscopic circumstances. For instance, the marker may cover some parts of the surgical ROI that can hamper the surgical workflow. Moreover, current design of the marker is single-sided and designed to be seen from the top, which makes the marker might not be always visible from the endoscopic camera when it is hidden behind the surgical instruments or out of the FOV of the camera. Note that simply reducing the dimension of the marker can't fundamentally resolve these limitations, nor secure the estimation accuracy. Hence, a new type of marker with a compact form-factor that fits into the laparoscopic circumstances is required. Recently, a cylindrically shaped marker was introduced which shows promising performance in the estimation accuracy when it is attached to the da Vinci surgical instrument [23]. Its robust performance yet compact design provided a potential utilization into other laparoscopic clinical applications. Therefore, a customized revision of the marker design tailored to our system will bolster the overall performance of the proposed algorithm.

Furthermore, an automated algorithm for efficient searching of the PM is under development within our research group. This will enable the system to become more efficient in the operating room. For example, manual manipulation for searching the PM in the TRUS domain will no longer required to the user, and it will ultimately shorten the registration time. In addition, the automated system will also enable an online re-calibration during the operation if it is needed. In particular, the position shift of the endoscopic camera might happen based on the surgeon's intention which requires frequent update of the registration. Here, the online re-calibration will enable the system to track the PM without any compromise.

In conclusion, we believe that the proposed arc-to-line based registration algorithm will take one step further towards the efficient integration of the US/PA imaging system into the clinics, especially contribute to the image-guided surgical intervention field.

## References

1. Lepor H (2005) A review of surgical techniques for radical prostatectomy. Rev Urology 7 Suppl 2:S11-7

2. GUILLONNEAU B, VALLANCIEN G (2000) LAPAROSCOPIC RADICAL PROSTATECTOMY: THE MONTSOURIS TECHNIQUE. J Urology 163:1643–1649. https://doi.org/10.1016/s0022-5347(05)67512-x

3. Ficarra V, Cavalleri S, Novara G, Aragona M, Artibani W, Villers A (2007) Evidence from Robot-Assisted Laparoscopic Radical Prostatectomy: A Systematic Review. Eur Urol 51:45–56. https://doi.org/10.1016/j.eururo.2006.06.017

4. Ficarra V, Novara G, Artibani W, Cestari A, Galfano A, Graefen M, Guazzoni G, Guillonneau B, Menon M, Montorsi F, Patel V, Rassweiler J, Poppel HV (2009) Retropubic, Laparoscopic, and Robot-Assisted Radical Prostatectomy: A Systematic Review and Cumulative Analysis of Comparative Studies. Eur Urol 55:1037–1063. https://doi.org/10.1016/j.eururo.2009.01.036

5. Dasgupta P, Kirby RS (2008) The current status of robot-assisted radical prostatectomy. Asian J Androl 11:90–93. https://doi.org/10.1038/aja.2008.11

6. Mitchell CR, Herrell SD (2014) Image-Guided Surgery and Emerging Molecular Imaging Advances to Complement Minimally Invasive Surgery. Urol Clin N Am 41:567–580. https://doi.org/10.1016/j.ucl.2014.07.011

7. Sridhar AN, Hughes-Hallett A, Mayer EK, Pratt PJ, Edwards PJ, Yang G-Z, Darzi AW, Vale JA (2013) Image-guided robotic interventions for prostate cancer. Nat Rev Urol 10:452–462. https://doi.org/10.1038/nrurol.2013.129

8. Raskolnikov D, George AK, Rais-Bahrami S, Turkbey B, Siddiqui MM, Shakir NA, Okoro C, Rothwax JT, Walton-Diaz A, Sankineni S, Su D, Stamatakis L, Merino MJ, Choyke PL, Wood BJ, Pinto PA (2015) The Role of Magnetic Resonance Image Guided Prostate Biopsy in Stratifying Men for Risk of Extracapsular Extension at Radical Prostatectomy. J Urology 194:105–111. https://doi.org/10.1016/j.juro.2015.01.072

9. Ukimura O, Magi-Galluzzi C, Gill IS (2006) Real-Time Transrectal Ultrasound Guidance During Laparoscopic Radical Prostatectomy: Impact on Surgical Margins. J Urology 175:1304–1310. https://doi.org/10.1016/s0022-5347(05)00688-9

10. Mohareri O, Ischia J, Black PC, Schneider C, Lobo J, Goldenberg L, Salcudean SE (2015) Intraoperative Registered Transrectal Ultrasound Guidance for Robot-Assisted Laparoscopic Radical Prostatectomy. J Urology 193:302–312. https://doi.org/10.1016/j.juro.2014.05.124

11. Long J-A, Lee BH, Guillotreau J, Autorino R, Laydner H, Yakoubi R, Rizkala E, Stein RJ, Kaouk JH, Haber G-P (2012) Real-Time Robotic Transrectal Ultrasound Navigation During Robotic Radical Prostatectomy: Initial Clinical Experience. Urology 80:608–613. https://doi.org/10.1016/j.urology.2012.02.081

12. Hung AJ, Abreu ALDC, Shoji S, Goh AC, Berger AK, Desai MM, Aron M, Gill IS, Ukimura O (2012) Robotic Transrectal Ultrasonography During Robot-Assisted Radical Prostatectomy. Eur Urol 62:341–348. https://doi.org/10.1016/j.eururo.2012.04.032

13. Bell MAL, Kuo NP, Song DY, Kang JU, Boctor EM (2014) In vivo visualization of prostate brachytherapy seeds with photoacoustic imaging. J Biomed Opt 19:126011–126011. https://doi.org/10.1117/1.jbo.19.12.126011

14. Zhang HK, Chen Y, Kang J, Lisok A, Minn I, Pomper MG, Boctor EM (2018) Prostate-specific membrane antigen-targeted photoacoustic imaging of prostate cancer in vivo. J Biophotonics 11:e201800021. https://doi.org/10.1002/jbio.201800021

15. Agarwal A, Huang SW, O'Donnell M, Day KC, Day M, Kotov N, Ashkenazi S (2007) Targeted gold nanorod contrast agent for prostate cancer detection by photoacoustic imaging. J Appl Phys 102:064701. https://doi.org/10.1063/1.2777127

16. Moradi H, Tang S, Salcudean SE (2019) Toward Intra-Operative Prostate Photoacoustic Imaging: Configuration Evaluation and Implementation Using the da Vinci Research Kit. Ieee T Med Imaging 38:57–68. https://doi.org/10.1109/tmi.2018.2855166

17. Cheng A, Zhang HK, Kang JU, Taylor RH, Boctor EM (2017) Localization of subsurface photoacoustic fiducials for intraoperative guidance. 100541D-100541D–6. https://doi.org/10.1117/12.2253097

18. Song H, Moradi H, Jiang B, Xu K, Wu Y, Taylor RH, Deguet A, Kang JU, Salcudean SE, Boctor EM (2022) Real-time intraoperative surgical guidance system in the da Vinci surgical robot based on transrectal ultrasound/photoacoustic imaging with photoacoustic markers: an *ex vivo* demonstration. Ieee Robotics Automation Lett PP:1–8. https://doi.org/10.1109/lra.2022.3191788

19. Visser M, Petr J, Müller DMJ, Eijgelaar RS, Hendriks EJ, Witte M, Barkhof F, Herk M van, Mutsaerts HJMM, Vrenken H, Munck JC de, Hamer PCDW (2020) Accurate MR Image Registration to Anatomical Reference Space for Diffuse Glioma. Front Neurosci-switz 14:585. https://doi.org/10.3389/fnins.2020.00585

20. Latifi K, Caudell J, Zhang G, Hunt D, Moros EG, Feygelman V (2018) Practical quantification of image registration accuracy following the AAPM TG-132 report framework. J Appl Clin Med Phys 19:125–133. https://doi.org/10.1002/acm2.12348

21. Pentenrieder K, Meier P, Klinker G (2006) Analysis of tracking accuracy for single-camera square-marker-based tracking

22. Song H, Kang J, Boctor EM (2023) Synthetic radial aperture focusing to regulate manual volumetric scanning for economic transrectal ultrasound imaging. Ultrasonics 129:106908. https://doi.org/10.1016/j.ultras.2022.106908

23. Cartucho J, Wang C, Huang B, Elson DS, Darzi A, Giannarou S (2022) An enhanced marker pattern that achieves improved accuracy in surgical tool tracking. Comput Methods Biomechanics Biomed Eng Imaging Vis 10:400–408. https://doi.org/10.1080/21681163.2021.1997647